\documentclass[10pt,twocolumn,letterpaper]{article}

\usepackage[pagenumbers]{cvpr} 

\usepackage{enumitem}

\definecolor{cvprblue}{rgb}{0.21,0.49,0.74}
\definecolor{qwen}{HTML}{7654ED}
\definecolor{darkred}{HTML}{8B0000}
\definecolor{lightcoral}{HTML}{F08080}
\definecolor{gaingreen}{HTML}{006400}
\usepackage[pagebackref,breaklinks,colorlinks,allcolors=cvprblue]{hyperref}
\usepackage{multirow}
\usepackage[table]{xcolor}
\usepackage{subcaption}
\usepackage{listings}
\usepackage{tcolorbox}
\usepackage{amssymb}
\usepackage{pifont}
\usepackage[T1]{fontenc}
\usepackage{marvosym}
\usepackage{algorithm,algpseudocode}
\newcommand{\cmark}{\ding{51}}%
\newcommand{\xmark}{\ding{55}}%

\newcommand{\ModelName}{MLLM-Sampler~Joint~Evolution}
\newcommand{\ModelAbbr}{MSJoE}
\newcommand{\DatasetNameAll}{LongVideoQA-ALL}
\newcommand{\DatasetNameHard}{LongVideoQA-HARD}

\def\paperID{} 
\def\confName{CVPR}
\def\confYear{2026}

\title{MSJoE: Jointly Evolving MLLM and Sampler for Efficient\\ Long-Form Video Understanding}

\author{
Wenhui Tan$^{1}$\thanks{This work was performed when Wenhui Tan was visiting Xiaomi as a research intern.} \quad
Xiaoyi Yu$^{2}$ \quad
Jiaze Li$^{3}$ \quad
Yijing Chen$^{1}$ \\
Jianzhong Ju$^{3}$ \quad
Zhenbo Luo$^{3}$ \quad
Ruihua Song$^{1}$\textsuperscript{\Letter} \quad
Jian Luan$^{3}$\textsuperscript{\Letter} \\
\\
$^{1}$Gaoling School of Artificial Intelligence, Renmin University of China \\
$^{2}$Tongji University \quad
$^{3}$MiLM Plus, Xiaomi Inc. \\
}

\begin{document}
\maketitle

\begin{abstract}
Efficiently understanding long-form videos remains a fundamental challenge for multimodal large language models (MLLMs).
In this paper, we present \ModelName~(\ModelAbbr), a novel framework that jointly evolves the MLLM and a lightweight key-frame sampler for efficient long-form video understanding.
\ModelAbbr~builds upon a key assumption that only a small subset of key-frames is truly informative for answering each question to a video.
Specifically, \ModelAbbr~first reasons out several queries, which describe diverse visual perspectives relevant to the question.
Then, these queries interact with a frozen CLIP model to produce a query–frame similarity matrix. Finally, A lightweight sampler predicts key-frame sampling weights from this matrix, selecting a compact set of informative frames, which are then fed into the MLLM for answer generation.
Both the MLLM and sampler are \textbf{jointly optimized through reinforcement learning}, enabling co-adaptation of query-reasoning, frame-sampling, and key-frame understanding.
A new long-video QA dataset containing 2.8k videos with 7k question–answer pairs is collected to support the training process.
Extensive experiments on VideoMME, LongVideoBench, LVBench, and MLVU show that \ModelAbbr~achieves 8.0\% accuracy gain upon the base MLLM, and 1.1\% higher accuracy than strongest baseline method.
\end{abstract}
\section{Introduction}\label{sec:intro}

\begin{figure}[t]
    \centering
    \includegraphics[width=1\columnwidth]{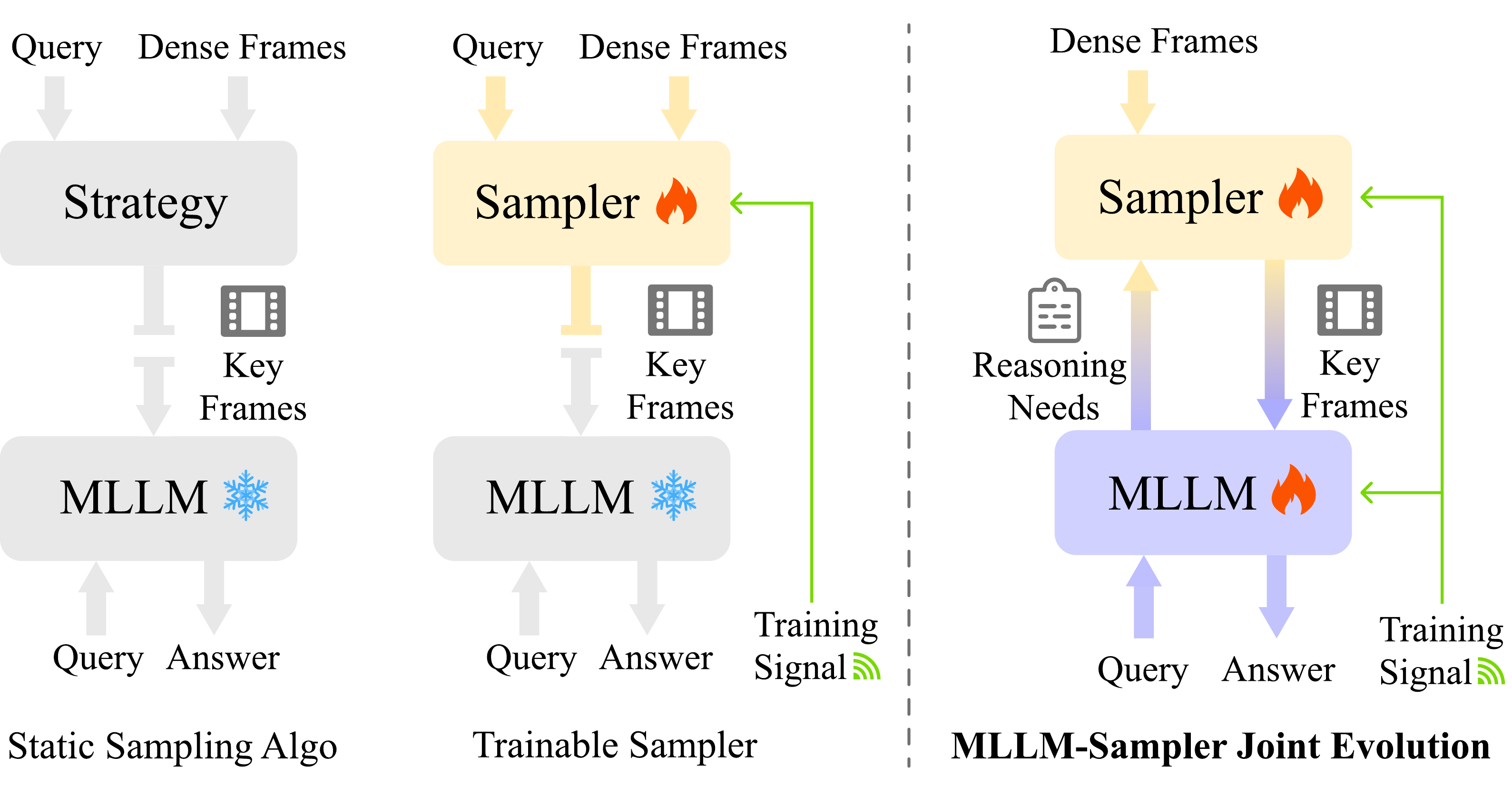}
    \caption{A direct comparison among static key-frame sampling algorithms, trainable key-frame sampler, and our proposed \ModelName~framework (\ModelAbbr).
    }
\end{figure}

Recent progress in multimodal large language models (MLLMs) has enabled strong performance in video understanding tasks such as captioning, reasoning, and question answering~\cite{gpt4o,gemini,qwen25vl,videollama,longvila}.
However, when the video gets longer, their efficiency and accuracy degrade rapidly: The visual context length scales linearly with duration, while attention computation grows quadratically, making traditional dense uniform sampling inefficient.
Furthermore, a question could involve multiple events in a video, while dense uniform sampling strategy is unlovable to overlook key events.

The core challenge lies in efficiently selecting informative frames from long-form videos, where most frames are visually similar or irrelevant to the question. Thus a fixed uniform-sampling frame budget forces the model to either miss key events or spend computation on uninformative regions. This motivates the key assumption of this work: \emph{only a small subset of frames (key-frames) is truly needed to answer a question about a long-form video}. Based on this assumption, we identify the fundamental question: \textbf{How to obtain the key-frames?}

To address this, many existing approaches leverage CLIP-based similarity between the \textit{question and frames} to locate relevant segments~\cite{tang2025aks,qframe,tspo2025,bolt}. However, this raises two further challenges:\\
\textbf{Q1: Is the question itself sufficient to retrieve all relevant frames?} (\emph{Insufficiency})\\
\textbf{Q2: How to effectively sample frames based on similarity scores?} (\emph{Sampling})

Often, the question lacks explicit visual cues, making CLIP retrieval unreliable. Thus, the question is needed to be decomposed. Furthermore, frame-wise similarity scores are not equivalent to key-frame sampling weights: a naive top-$k$ strategy over similarity scores tends to select redundant frames. Hence, some works proposed heuristic algorithms~\cite{bolt,tang2025aks,qframe} to transform similarity scores into key-frame sampling weights. However, these methods often require careful algorithm design or even specific tuning on different datasets. TSPO~\cite{tspo2025} proposes a trainable sampler that learns to select frames from CLIP similarities, yet it overlooks the fact: most MLLMs are pre-trained on uniformly sampled videos instead of key-frames~\cite{qwen25vl,videollama,llavavideo,llavaov}.
Hence, we raise the last question:\\
\textbf{Q3: Can the MLLM and sampler truly collaborate without joint evolution?} (\emph{Collaboration})

Effective collaboration requires two capabilities: (i) the MLLM must learn to generate reasoning queries that guide keyframe selection, and (ii) the MLLM must adapt to reason over the sparse keyframes that the sampler provides. Current methods freeze the MLLM during sampler training, preventing this bidirectional adaptation.

To address these issues, we propose an MLLM-Sampler Joint Evolution framework (\ModelAbbr) for efficient long-form video understanding.
To address the \emph{insufficiency} (\textbf{Q1}) of information to retrieve frames with question, \ModelAbbr~first reason out potential helpful perspectives, generating multiple queries that describe visual events or clues relevant to answering the question.
These queries are paired with densely sampled frames to form a query–frame similarity matrix via a frozen CLIP model. A lightweight 1D U-Net sampler with about two million parameters \emph{learns} (\textbf{Q2}) to transforms this matrix into sampling weights, thus samples a small set of diverse and informative key-frames.
Finally, the selected frames are fed back into the MLLM, which is then trained jointly with the sampler through reinforcement learning (RL). This end-to-end optimization ensures \emph{aligned progress} (\textbf{Q3}) in frame selection and understanding.

Due to lack of multi-hop and hour-long video datasets, we newly collected a long-video QA dataset, and we apply reinforcement learning on Qwen2.5-VL-7B-Instruct~\cite{qwen25vl} as our proposed \ModelAbbr. Experiments on VideoMME, LongVideoBench, LVBench, and MLVU~\cite{videomme,longvideobench,lvbench,mlvu} show that \ModelAbbr~outperforms existing sampling-based and long-context MLLMs, achieving a superior balance between computational cost and accuracy.

Our main contributions are as follows:
\begin{itemize}
    \item We propose \textbf{\ModelName~(\ModelAbbr)}, a unified framework that jointly evolves an MLLM and a trainable sampler, allowing reasoning-guided key-frame selection and co-adaptation of perception and language understanding.
    \item We introduce a new long-video QA dataset with 2.8k videos and 7.1k question-answer pairs, supporting reinforcement learning for joint optimization.
    \item Extensive experiments on VideoMME, LongVideoBench, LVBench, and MLVU show that \ModelAbbr~achieves 8.0\% accuracy gain upon the base MLLM, and 1.1\% higher accuracy than strongest baseline method.
\end{itemize}

\section{Related Work}
\subsection{Long-Form Video Understanding Methods with Key Frame Sampling}

Recent advances in Multimodal Large Language Models (MLLMs) have enabled remarkable progress in video understanding tasks~\cite{gpt4o,gemini,qwen25vl,videollama,longvila}. However, processing long-form videos remains constrained by limited context windows~\cite{focus,DATE}. Existing approaches fall into two categories: training-free methods and trainable sampling strategies.

\textbf{Training-Free Sampling Methods.}  Uniform sampling methods like LLaVA-Video~\cite{llavavideo}, Qwen2.5-VL~\cite{qwen25vl} and SlowFast-LLaVA~\cite{xu2024slowfastllava} rely on the assumption of a uniform temporal distribution. LongVU~\cite{shen2024longvu} shows this approach naively overlooks non-uniform content and misses critical frames in sparse event scenarios. To address uniform sampling's blindness to content semantics, CLIP-based methods attempt content-aware selection but face the insufficiency problem (\textbf{Q1}): questions in video QAs are often interrogative and temporally focused rather than descriptive, providing insufficient visual grounding. They specify what information is needed without describing where or how it appears visually. Despite this limitation, methods like VaQuitA~\cite{wang2023vaquita}, AKS~\cite{tang2025aks} and QFrame~\cite{qframe} employ similarity-based retrieval and address \textbf{Q2} through heuristic algorithms that transform similarity scores into sampling weights. However, they require carefully designed test-time strategies that do not generalize well to diverse video-query pairs.

\begin{figure*}[t]
\centering
\includegraphics[width=1\textwidth]{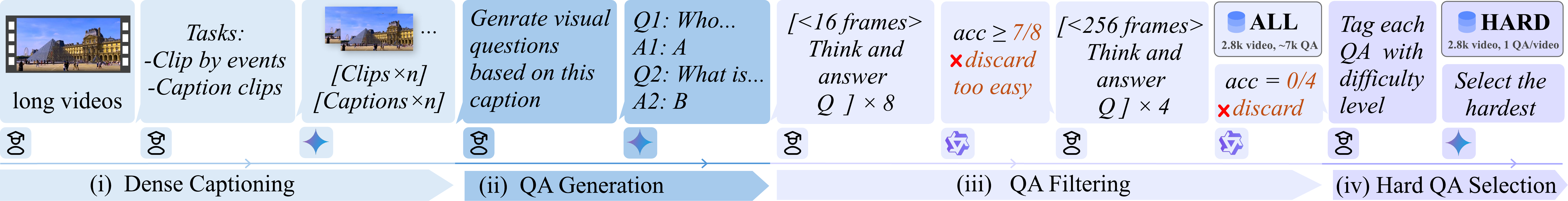}  
\caption{Our dataset construction pipeline. QA pairs with low difficulty or poor quality are removed during multi-stage filtering.}
\label{fig:dataset}
\end{figure*}

\textbf{Trainable Sampling Strategies.} To overcome the limitations of fixed heuristics, learning-based methods aim to solve \textbf{Q2} by training samplers to predict key-frame weights. TSPO~\cite{tspo2025} employs GRPO to train a lightweight temporal agent. K-frames~\cite{yao2025kframes} uses a smaller MLLM to predict key clips for a larger MLLM. M-LLM Frame Selector~\cite{hu2025mllm} trains a learnable projector to predict frame importance. ViaRL~\cite{viarl} alternates between training the selector and fine-tuning the MLLM.

However, these methods share an architectural constraint: the MLLM remains frozen during sampler training or is optimized separately.  This design choice prevents \emph{collaboration} (\textbf{Q3}): the MLLM cannot learn to generate effective reasoning queries for guiding frame selection, nor can it adapt to reason over the sparse keyframes the sampler provides. The sampler receives only sparse scalar rewards without understanding which visual evidence the MLLM needs, while the MLLM remains optimized for uniform sampling rather than the keyframe distributions its sampler produces.

In summary, existing methods progress on \emph{Sampling} (\textbf{Q2}), but question \emph{insufficiency} (\textbf{Q1}) remains underexplored, and the lack of \emph{collaboration} (\textbf{Q3}) represents a fundamental limitation. Without joint evolution, neither component can adapt to the other: the MLLM cannot guide keyframe selection through reasoning queries, nor can it learn to effectively process the sparse visual inputs the sampler provides.

\subsection{Training MLLMs with RL}

Reinforcement Learning has evolved from aligning language models with human preferences to enabling sophisticated multimodal reasoning capabilities. RLHF~\cite{ouyang2022training} established the foundation by using human feedback to train reward models, treating LLMs as policy networks optimized through PPO~\cite{schulman2017proximal}. DPO~\cite{rafailov2023direct} simplified this pipeline by eliminating explicit reward models, directly optimizing from preference pairs. Recent multimodal extensions include LLaVA-RLHF~\cite{sun2023aligning}, which introduced factually augmented rewards to reduce hallucinations by 39.5\%, and RLHF-V~\cite{yu2024rlhf}, which achieved 75.8\% hallucination reduction through dense segment-level corrections. DeepSeek-R1~\cite{deepseekr1} demonstrated that GRPO~\cite{deepseekmath} can induce emergent reasoning without supervised fine-tuning. For long-form video understanding, TSPO~\cite{tspo2025} applies GRPO to learn adaptive frame sampling policies, but their approach relies on a fixed-query to guide keyframe selection while freezing the LLM during training, hindering the collaboration and joint evolution of the sampler and the MLLM components.
\section{Dataset}\label{sec:data}

A major bottleneck in training long-form video understanding models lies in the lack of large-scale datasets with sufficiently long durations and reasoning-oriented questions. To train \ModelName, we construct a new long-video question answering dataset designed to satisfy four goals: (i) videos with long durations and rich temporal structure, (ii) questions that require reasoning across multiple events, (iii) an efficient and automatic pipeline for scalable data generation, and (iv) controllable question difficulty for reinforcement learning.
The overall pipeline consists of four steps: Step 1: Dense Captioning, 
Step 2: QA Generation, Step 3: QA Filtering, and Step 4: Difficulty Labeling and Subset Selection.
The dataset construction pipeline is illustrated in Figure~\ref{fig:dataset}.
Due to space limit, we elaborate the data collection pipeline in the Supplementary Material Section~\ref{sup:dataset}.
\section{Method}

\begin{figure*}[t]
\centering
\includegraphics[width=1\textwidth]{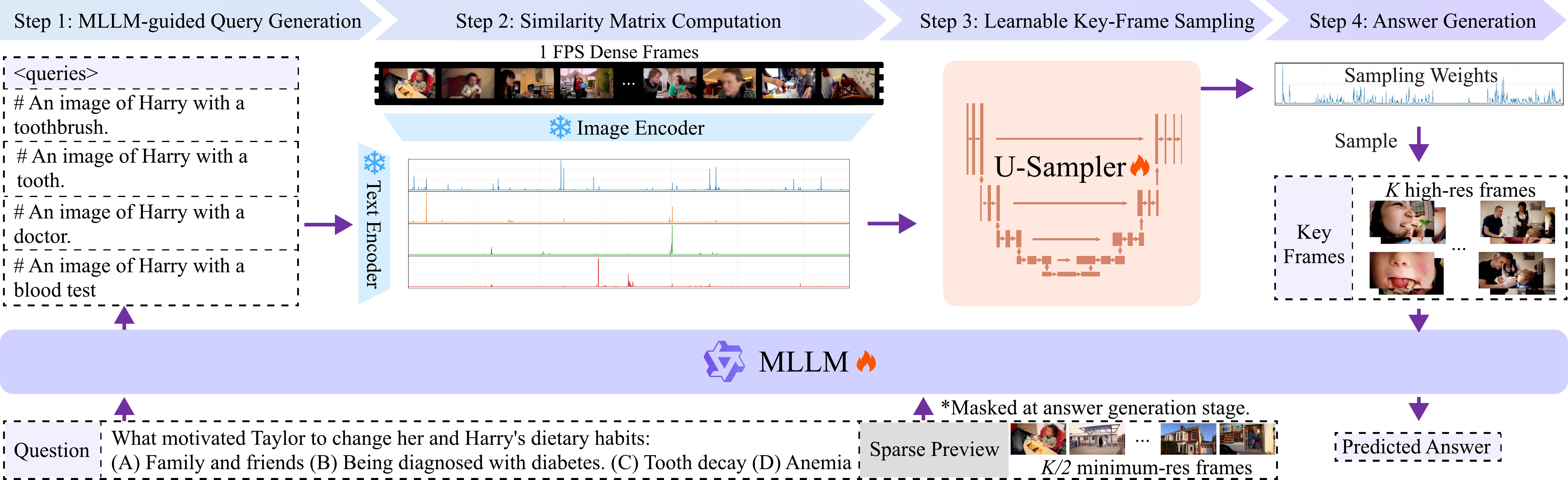}  
\caption{The proposed \ModelAbbr~ framework. Given a video and question, \ModelAbbr~generates reasoning-based queries from a sparse preview, matches them against dense frames via CLIP to create a similarity matrix, and uses a lightweight U-Sampler to select informative frames. The MLLM then processes these key frames at high resolution for answer generation. The entire framework is jointly optimized through end-to-end reinforcement learning.}
\label{fig:main}
\end{figure*}


\subsection{Overview}
Given a long video $\mathcal{V}$ with $T$ frames and a question $q$, our goal is to select a subset of $K$ informative frames ($K \ll T$) that preserves the precision of the answer while maintaining computational efficiency. Our method \ModelAbbr~achieves this by enabling collaboration between the MLLM and a lightweight sampler. In this section, we will illustrate the inference and training pipeline in Section~\ref{sec:method:inference} and Section~\ref{sec:method:train}, respectively.

\subsection{Inference Pipeline}\label{sec:method:inference}
As illustrated in Figure~\ref{fig:main}, the pipeline consists of four steps: (1) MLLM-guided Query Generation that reasons out informative visual queries based on question and a sparse video preview, (2) CLIP-based Similarity Computation to obtain query-frame similarity distribution, (3) Learnable Key-Frame Sampling that adaptively learns to sample key-frames from the similarity matrix, and (4) Answer Generation that reasons over the sampled key-frames to produce the final answer. Note that the sparse video preview is masked/removed during answer generation for fair comparison to non-reasoning methods.

\paragraph{Step 1: MLLM-guided Query Generation.}
We begin by uniformly sampling a sparse set of $N_{\text{init}}$ frames from the video at minimum resolution (only 32 tokens per frame). These frames provide a coarse preview of the video content without introducing much overhead. The MLLM processes the question $q$ along with these preview frames to reason out $N_q$ queries $\{q^1, q^2, \ldots, q^{N_q}\}$. These queries describe specific visual patterns or events that would be helpful for answering the question.

Formally, let $\mathcal{F}_{\text{init}} = \{f_i\}_{i=1}^{N_{\text{init}}}$ denote the initial sparse frames. The MLLM generates queries as:
\begin{equation}
\{q^j\}_{j=1}^{N_q} = \text{MLLM}(q, \mathcal{F}_{\text{init}}; \theta_{\text{MLLM}})
\end{equation}
where $\theta_{\text{MLLM}}$ represents the MLLM parameters.

\paragraph{Step 2: Similarity Matrix Computation.}
We densely sample the video at 1 FPS to obtain $N_f$ frames. Each generated query $q^j$ and frame $f_i$ are encoded using a frozen CLIP model:
\begin{equation}
s_{ji} = \text{sim}(\text{CLIP}_{\text{text}}(q^j), \text{CLIP}_{\text{image}}(f_i))
\end{equation}
where $\text{sim}(\cdot, \cdot)$ denotes cosine similarity. This produces a similarity matrix $\mathbf{S} \in \mathbb{R}^{N_q \times N_f}$, where each row represents how well a specific query matches across all frames.

\paragraph{Step 3: Learnable Key-Frame Sampler.}
The sampler with parameter $\phi$ takes the similarity matrix $\mathbf{S}$ as input and produces per-frame sampling probabilities $\mathbf{p}\in\mathbb{R}^{N_f}$.
Considering the nature of $\phi: \mathbf{S}\in\mathbb{R}^{N_q \times N_f} \to \mathbf{p} \in\mathbb{R}^{N_f}$: \textbf{dense-prediction} and \textbf{local-perception}, the sampler is implemented with an 1D U-Net~\cite{unet}, treating the query axis as channles.

Aligning with previous works, we sample $K$ frames with indices $\textbf{x}=\{x\}_{i=1}^K$ according to distribution $\textbf{p}$, and each selected frame is encoded at normal resolution (256 tokens per frame) for final answer generation.

\paragraph{Step 4: Answer Generation.}
The MLLM processes the selected frames $\mathcal{F}_{\text{selected}}$ along with the original question to generate the final answer:
\begin{equation}
a = \text{MLLM}(q, \mathcal{F}_{\text{selected}}; \theta_{\text{MLLM}})
\end{equation}

The conversation history, including the query generation step, is maintained to provide context for answer generation, while the sparse preview is masked. This allows the MLLM to understand why certain frames were selected based on its own generated queries.

\subsection{Training Pipeline}\label{sec:method:train}

\subsubsection{Joint RL Training}
The MLLM and sampler of \ModelAbbr~is trained end-to-end with RL, which enables joint optimization of both the MLLM and sampler components. This subsection elaborates the reward and objective of the RL training process.

\paragraph{Reward Design.}
The total reward $r$ for a training sample consists of three components:

\textbf{Accuracy reward} $r_{\text{acc}}$: We assign a reward of 0.8 if the generated answer is correct, and 0 otherwise. This provides the primary supervision signal for both components.


\textbf{Format Reward} $r_{\text{format}}$: The model is encouraged to produce a well-formatted response, from query reasoning to question answering. We assign a reward of 0.1 if the model produces a response in correct format.

\textbf{Informativeness reward} $r_{\text{info}}$: We encourage queries that produce peaked similarity distributions with clear high-attention regions:
\begin{equation}
r_{\text{info}} = 0.1 \cdot \frac{\sum_{i} \mathbb{I}\left[\frac{\max_i s_{ji}}{\min_i s_{ji}} > \tau_{\text{info}} \right]}{N_q}
\end{equation}
where $\tau_{\text{info}}$ is an informativeness threshold. This punishes general queries that match all frames equally.

\paragraph{Training Objectives.}
We use GRPO algorithm~\cite{deepseekmath} to train the MLLM: for each training question $q$, we sample a group of $G$ outputs $\{o\}_{i=1}^{G}$ by generating different query sets and frame selections. Then the MLLM with parameter $\theta$ is optimized with GRPO objective:
\begin{align}
    \mathcal{J}_{\text{G}}(\theta) & = \mathbb{E}_{o_1, \dots, o_G \sim \pi_{\theta_{\text{old}}}, q \sim \mathrm{Q}}\\
    &\left[\frac{1}{G}\sum_{i=1}^{G}\left(
        \min \left(
            s_iA_i,
            \text{clip}\left(
                s_i,
                1 - \epsilon,
                1 + \epsilon
            \right) A_i
        \right)
    \right)\right],
\end{align}
where
\begin{equation}
s_i=\frac{\pi_{\theta}\left(o_i | q \right)}{\pi_{\theta_{\text{old}}}\left(o_i | q \right)},
\end{equation}
denoting importance sampling, and 
\begin{equation}
    A_i = \frac{
        r_i - \text{mean}\left( {r_1, r_2, \ldots, r_G} \right)
    }{
        \text{std}\left( {r_1, r_2, \ldots, r_G} \right)
    },
\end{equation} 
denoting group-relative advantage. The reward for each output is 
$r_{\text{G}} = r_{\text{acc}} + r_{\text{format}} + r_{\text{info}}$.

The sampler with parameter $\phi$ is trained with REINFORCE objective:
\begin{equation}\label{eq:sampler}
    \mathcal{J}_{\text{R}}(\phi) = \mathbb{E}_{x_1, \dots, x_K \sim \mathbf{p}} 
    \left[ A_{\text{sampler}} \cdot \sum_{k=1}^{K} \nabla_{\phi} \log \mathbf{p}(x_k) \right],
\end{equation}
where $A_{\text{sampler}}$ shares the same reward of accuracy $r_{\text{acc}}$.

Through this joint optimization, the MLLM learns to generate queries that facilitate frame retrieval and understanding, while the sampler learns to select frames that maximize reasoning effectiveness.

\subsubsection{Sampler Pre-Training}
Though the RL process enables joint evolution of both MLLM and sampler, a randomly initialized sampler can introduce excessive noise in the early stage of joint training. To stabilize learning, we propose an auxiliary pre-training stage to the sampler on \DatasetNameAll~before entering the joint RL phase. The training objective is the same to Eq.~\ref{eq:sampler}.

GRPO algorithm could also be used for this pre-training phase. However, the computational cost is about $G\times$ higher than REINFORCE algorithm. On the other hand, without group-level optimization, a standalone REINFORCE setup with a binary reward can be misleading in our setting: when the correct key frames are selected, but the answer remains wrong due to high question difficulty, the binary reward still penalizes the sampler, which discourages it from trusting key frames. 

To mitigate this issue, we introduce a \emph{difficulty-aware reward}. As described in Section~\ref{sec:data}, each QA pair in \DatasetNameAll~has an associated pass rate $c \in [0,c)$, estimated by the MLLM’s accuracy under uniform frame sampling. This rate reflects the inherent difficulty of the question.

When $c=0$, which means that the base MLLM never answers the question correctly even once with uniform sampling, we treat any correct answering during sampler pre-training as a strong signal of key-frame discovery, and assign a high reward with $A_{\text{sampler}}=10$; while an incorrect answer receive $A_{\text{sampler}}=0$, implying no penalty for extremely difficult questions.
When $c \neq 0$, the reward is defined as $A_{\text{sampler}}=\frac{1}{c}$ for correct answers and $A_{\text{sampler}}=\frac{-1}{\left(1-c\right)}$ for incorrect answers (there is no zero-division risk as all-pass questions are filtered out). This encourages the sampler to prioritize harder questions and penalizes failures on easier ones.

This difficulty-aware reward provides smoother and more informative gradients for pre-training, allowing the sampler to focus on discovering frames that meaningfully improve the MLLM’s performance.

\section{Experiments}\label{sec:exp}

\begin{table*}[t]
\centering
\caption{Comparison of MLLMs and baseline methods against our method~\ModelAbbr~on four benchmarks. We bold the best results and highlight performance gain over the base-MLLM. The performance of baseline methods are reported according to published papers.}
\label{tab:sota}
\begin{tabular}{lcccccc}
\hline
\multirow{2}{*}{Method} & \multirow{2}{*}{\#Frames} & \multirow{2}{*}{MLVU} & \multirow{2}{*}{LongVideoBench} & \multicolumn{2}{c}{Video-MME} & \multirow{2}{*}{LVBench} \\
                        &                           &                                 &                       & Long               & Avg.     &                          \\ \toprule
\rowcolor{gray!50}
\multicolumn{7}{l}{\textit{\textbf{Close-Sourse MLLMs}}}                                                                                                                 \\
GPT-4o                  & -                         & 66.7                            & 64.6                  & 65.3               & 71.9     & -                        \\
Gemini-1.5-pro          & -                         & 64.0                            & -                     & 67.4               & 75.0     & 33.1                     \\ \midrule
\rowcolor{gray!10}
\multicolumn{7}{l}{\textit{\textbf{Open-Source MLLMs}}}                                                                                                                  \\
VideoMind-7B            & -                         & -                               & 64.4                  & 49.2               & 58.2     & 40.8                     \\
LongVU-7B                  & 1 FPS                      & -                               & 65.4                  & -                  & 60.6     & -                        \\
NVILA-8B                   & 1024                      & 57.7                            & 70.1                  & 54.8               & 64.2     & -                        \\
Qwen-2.5-VL-7B          & 768/2 FPS                  & 56.0                            & 70.2                  & -                  & 65.1     & 45.3                     \\ \midrule
\rowcolor{gray!10}
\multicolumn{7}{l}{\textit{\textbf{Qwen-2.5-VL-7B based methods (32 frames)}}}                                                                                           \\
Uniform Sampling        & 32                        & 61.5                  & 55.0                            & 49.9               & 63.7     & 36.5                     \\
Q-Frame                 & 32                        & 66.8                  & 58.7                            & 53.1               & 62.6     & -                        \\
BOLT                    & 32                        & 66.3                  & 58.6                            & 53.8               & 64.1     & -                        \\
\rowcolor{qwen!10}
MSJoE (Ours)           & 32                        & \textbf{69.3} \small{\textcolor{gaingreen}{(+7.8)}}    & \textbf{60.1} \small{\textcolor{gaingreen}{(+5.1)}}                  & \textbf{54.1} \small{\textcolor{gaingreen}{(+4.2)}}     & \textbf{64.3}        & \textbf{46.4} \small{\textcolor{gaingreen}{(+9.9)}}           \\ \hline
\rowcolor{gray!10}
\multicolumn{7}{l}{\textit{\textbf{Qwen-2.5-VL-7B based methods (64 frames)}}}                                                                                           \\
Uniform Sampling        & 64                        & 65.3                  & 57.3                            & 52.2               & 64.1     & 39.2                     \\
TSPO                    & 64                        & 74.3                  & \textbf{64.2}                   & 56.4               & 65.5     & 46.4                     \\
\rowcolor{qwen!10}
MSJoE (Ours)           & 64                        & \textbf{75.1} \small{\textcolor{gaingreen}{(+9.8)}}         & 62.2  \small{\textcolor{gaingreen}{(+4.9)}}                          & \textbf{57.4} \small{\textcolor{gaingreen}{(+5.2)}}     & \textbf{66.2}        & \textbf{51.1} \small{\textcolor{gaingreen}{(+11.9)}}           \\ \hline
\end{tabular}
\end{table*}

\subsection{Setup}\label{sec:exp:setup}
\paragraph{Benchmarks.}
We evaluate the efficacy of our proposed method \ModelAbbr~on four widely used long-form video understanding benchmarks:
\begin{itemize}
    \item \textbf{LongVideoBench}~\cite{longvideobench} consists of 1,337 QA pairs on 753 videos with an average duration of 12 min.
    \item \textbf{MLVU}~\cite{mlvu} consists of 2,174 QA pairs on 1,242 videos with an average duration of 12 min. 
    \item \textbf{VideoMME}~\cite{videomme} consists of three subsets classified by video duration (short: 1.3 min, medium: 9 min, long: 41 min, average: 17 min). Each subset consists of 300 videos with three questions for each video.
    \item \textbf{LVBench}~\cite{lvbench} consists of 1,549 QA pairs on 103 hour-long videos.
\end{itemize}
Note that the statistics provided are based on the Multichoice Question Answering (MQA) subset of the benchmarks, and the reported performance is the MQA accuracy.

\paragraph{Baseline Methods.}
We compare our proposed method \ModelAbbr~with State-of-the-Art (SOTA) methods, which could be classified into three main categories:
\begin{itemize}
    \item \textbf{Close-Source MLLMs}: We report the performance of cutting-edge proprietary models GPT-4o~\cite{gpt4o} and Gemini-1.5-pro~\cite{gemini}.
    \item \textbf{Open-Source MLLMs}: We compare our method to pretrained or finetuned MLLMs, including reasoning-based MLLM VideoMind, long-form video oriented MLLMs LongVU~\cite{longvu} and NVILA~\cite{nvila}, and base-MLLM Qwen-2.5-VL~\cite{qwen25vl}. 
    \item \textbf{Key-Frame Sampling Based Methods}: We directly compare our method with heuristic key-frame sampling methods Q-Frame~\cite{qframe} and BOLT~\cite{bolt}, and training-based key-frame sampling method TSPO~\cite{tspo2025}. For fair comparison, the performances are reported based on the same base-MLLM and frame budget.
\end{itemize}

\paragraph{Implementation Details.}
We build \ModelAbbr~on Qwen2.5-VL-7B-Instruct~\cite{qwen25vl} as the base MLLM and use a frozen Clip-ViT-Large-Patch14~\cite{radford2021clip} for similarity computation. The sampler is implemented with a U-Net~\cite{unet} architecture with 1-D convnets. The number of frames of the initial preview is set to $N_{\text{init}}=K//2$.
During reinforcement learning, we use a batch size of 32 and train for two epochs with a learning rate of 1e-6 for the MLLM and 1e-5 for the sampler. We set $\tau_\text{info} = 10$ as informativeness reward threshold. For each training sample, we generate $G=8$ trajectory samples for GRPO. The entire framework is mainly implemented in PyTorch~\cite{pytorch}, VERL~\cite{verl} and vLLM~\cite{vllm}.
For more implementation details, please refer to Supplementary Materials Section~\ref{sup:implementation}.

\subsection{Comparison to Baseline Methods}\label{sec:exp:sota}
As shown in Table~\ref{tab:sota}, our proposed method \ModelAbbr~achieves state-of-the-art performance compared to all open-source baselines. Several key observations can be drawn from these results:
\paragraph{\ModelAbbr~improves upon both the base MLLM and the strongest baselines.}
\ModelAbbr~consistently outperforms its base model, Qwen-2.5-VL-7B, across all benchmarks, achieving an average improvement of \textbf{+6.7 points} with a 32-frame sampling budget and \textbf{+8.0 points} with a 64-frame budget. Even when using only half of the frames, \ModelAbbr~still surpasses the base MLLM by \textbf{+4 points}. 
Compared to the strongest existing baseline, TSPO, our method achieves an additional \textbf{+1.1 points} improvement on average. 
These results demonstrate that \ModelAbbr~provides superior long-form video understanding, achieving higher accuracy while maintaining efficiency.
\paragraph{Key-frame sampling methods exhibit higher efficiency.}
Compared with dense-frame models such as LongVU and NVILA, which rely on uniformly or densely sampled video frames, key-frame sampling approaches (from Q-Frame to \ModelAbbr) achieve consistently better accuracy. These results support our key claim: \emph{only a small subset of frames (key-frames) is truly needed to answer a question about a long-form video}.
\paragraph{Learned sampling surpasses heuristic algorithms.}
When compared with the state-of-the-art heuristic key-frame sampler BOLT, \ModelAbbr~achieves a significant performance gain of \textbf{0.3-3.0 points} under identical frame budgets. TSPO also shows more improvement over its base MLLM, but \ModelAbbr~further extends this advantage.
These results suggest that: learning-based sampling strategies are more capable of identifying informative frames than heuristic approaches, leading to more effective video comprehension under constrained sampling conditions.

\begin{figure*}[h]
    \centering
    \begin{subfigure}{0.22\textwidth}
        \centering
        \includegraphics[width=\textwidth]{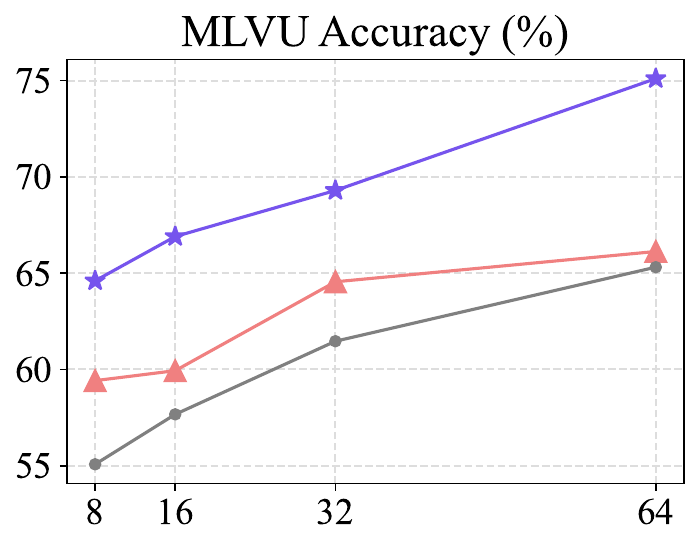}
    \end{subfigure}
    \hfill
    \begin{subfigure}{0.22\textwidth}
        \centering
        \includegraphics[width=\textwidth]{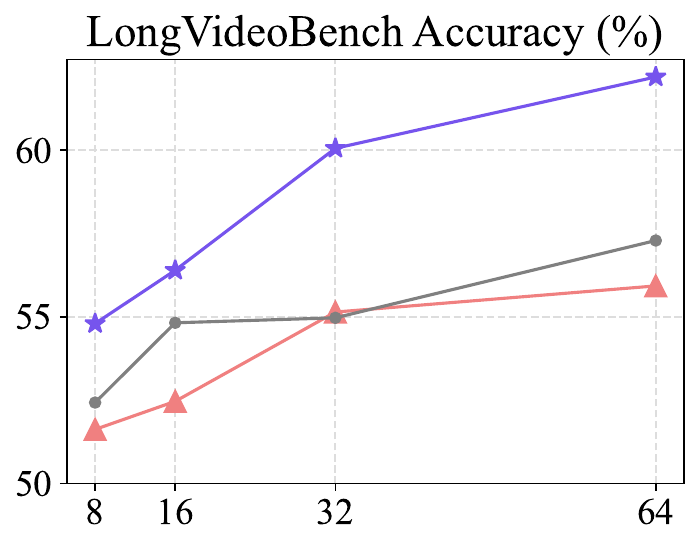}
    \end{subfigure}
    \hfill
    \begin{subfigure}{0.22\textwidth}
        \centering
        \includegraphics[width=\textwidth]{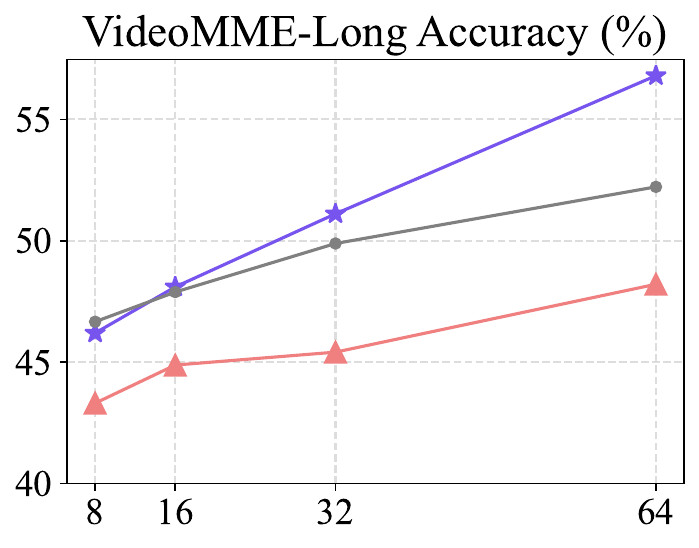}
    \end{subfigure}
    \hfill
    \begin{subfigure}{0.22\textwidth}
        \centering
        \includegraphics[width=\textwidth]{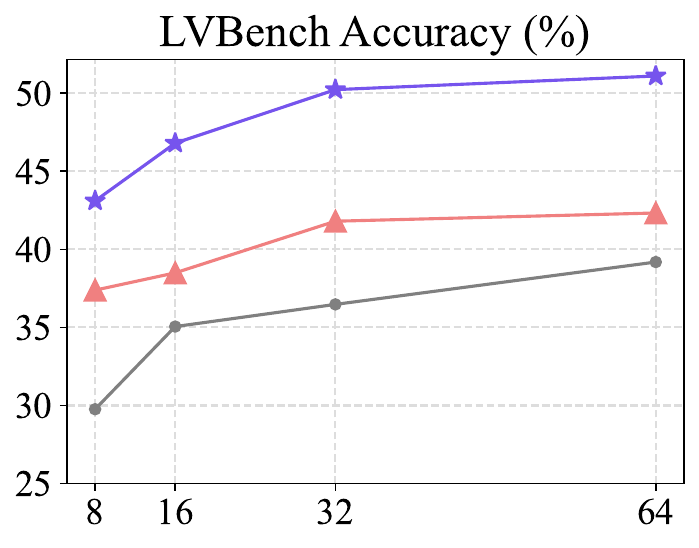}
    \end{subfigure}
    \caption{Ablation studies on varying input frames (x-axis). Four methods are evaluated: \textcolor{qwen}{MSJoE} in light violet, \textcolor{lightcoral}{Top-$k$} in red, and \textcolor{gray}{Uniform Sampling} uniform sampling in gray.}
    \label{fig:duration}
\end{figure*}

\begin{table}[h]
\centering
\caption{Ablation studies on the MLLM (M) and Sampler (S) module with a fixed input budget of 32 frames. The module settings T, PT, and F denote Training, Pre-Trained, Frozen respectively. Setting \emph{iii} (PT-T*) denotes we feed the same frames (reasoned and sampled) as \emph{Ours} to a frozen MLLM. Setting \emph{vi} (F*) denotes using a Frozen MLLM w/ question-as-query. LoVi, VLong, and LVB denote the benchmarks LongVideoBench, VideoMME-Long, and LVBench, respectively.}
\label{tab:abla:mllm_sampler}
\begin{tabular}{lcc|cccc}
\toprule
\multirow{2}{*}{} & \multirow{2}{*}{M} & \multirow{2}{*}{S} & \multirow{2}{*}{MLVU} & \multirow{2}{*}{LoVi} & \multirow{2}{*}{VLong} & \multirow{2}{*}{LVB} \\
                  &                    &                    &                       &                       &                        &                      \\ \midrule
\rowcolor{qwen!10}
Ours              & T                  & PT-T               & 69.3                  & 60.1                  & 54.1                   & 46.4                 \\
(i)               & T                  & T                  & 69.1                  & 59.8                  & 53.8                   & 46.2                 \\
(ii)              & T                  & PT-F               & 68.8                  & 58.1                  & 52.4                   & 45.5                 \\ \midrule
(iii)             & F                  & PT-T*              & 68.7                  & 58.2                  & 52.1                   & 44.8              \\
(iv)              & F                  & PT-F               & 68.2                  & 56.8                  & 52.2                   & 44.3             \\
(v)               & F                  & top-k              & 64.5                  & 55.1                  & 45.4                   & 41.8             \\
(vi)              & F*                 & top-k              & 68.3                  & 56.2                  & 48.9                   & 44.6             \\
\rowcolor{gray!10}
Uni               & F                  & uni                & 61.5                  & 55.0                  & 49.9                   & 36.5                 \\ \midrule
\end{tabular}
\end{table}

\subsection{Ablation Study}\label{sec:exp:ablation}
We evaluate the contribution of each component in \ModelAbbr~through controlled ablations across four benchmarks, as shown in Table~\ref{tab:abla:mllm_sampler}.

\subsubsection{Answering the Questions}
We now answer the questions raised in Section~\ref{sec:intro}.

Comparing setting \textbf{(vi)} and \textbf{uniform sampling}, we observe that directly using the question as a query yields a clear improvement, up to +3.8 average points over uniform sampling. This indicates that CLIP’s strong prior can align the question with helpful visual cues and retrieve more informative frames. However, due to the inherent gap between a linguistic query and a visual concept, its performance remains below the frozen-MLLM variants \textbf{(iii)} and \textbf{(iv)}.  
Therefore, \textbf{Q1: Is the question alone sufficient for retrieving all relevant frames? Answer: Not sufficient}.

To alleviate the insufficiency of Q1, we further decompose the question into multiple queries \textbf{(v)}. Yet, performance drops significantly compared to using the single question. We identify two causes:
\begin{itemize}
    \item An 1D similarity is easier to sample than a 2D similarity matrix. Applying top-$k$ over an $N_\text{query}\times N_\text{frames}$ matrix requires pooling, but both average- and weighted-pooling yield suboptimal results. Only when paired with a trained sampler \textbf{(iv)} does the multi-query strategy surpass vanilla top-$k$. Thus, \textbf{Q2: How to effectively sample frames from similarity scores? Answer: A well-trained sampler is necessary}.
    \item A frozen MLLM cannot produce strong clip-oriented queries without training. From the comparisons \textbf{(ii)$\rightarrow$(iv)} and \textbf{Ours$\rightarrow$(iii)}, a trained MLLM consistently outperforms a frozen one: it is trained to reason out better queries, and, even with identical frame inputs, understands them more effectively. Thus, \textbf{Q3: Can the MLLM and sampler truly collaborate without joint evolution? Answer: No. Co-evolution is required to improve reasoning and key-frame understanding}.
\end{itemize}

\subsubsection{Further Analyses}
We further analyze \ModelAbbr~to validate both its effectiveness and efficiency.

First, as shown in Table~\ref{tab:abla:mllm_sampler}, comparing our method with setting \textbf{(i)} indicates that a pre-trained sampler provides a stronger initialization for RL optimization, leading to consistently improved performance.

Second, Figure~\ref{fig:duration} shows that our approach consistently outperforms the uniform sampling baseline across various input budgets \{8, 16, 32, 64\} frames. Moreover, \ModelAbbr~achieves same or higher performance with much fewer number of frames, showing superior efficiency.
In contrast, the top-$k$ sampling strategy can even degrade performance on LongVideoBench and VideoMME-Long, suggesting that naive similarity-based selection is less stable.



\begin{figure*}[h]
    \centering
    \includegraphics[width=\textwidth]{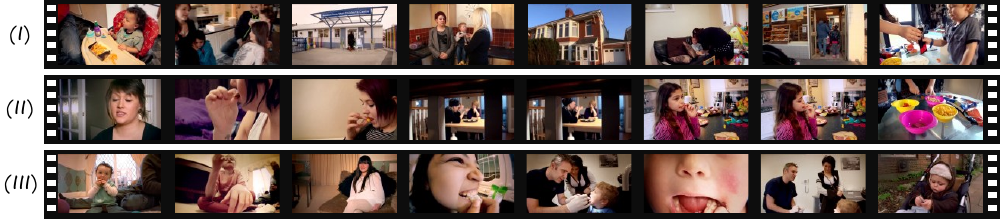}
    \caption{Three frame sets from a publicly available video generated by different sampling strategies. \emph{Question: What motivated her to change dietary habits? (A) Family and friends (B) Diabetes (C) Tooth decay (D) Anemia}.}
    \label{fig:case}
\end{figure*}

\subsection{Case Study}\label{sec:exp:case}
We present a qualitative case study to illustrate how different sampling strategies influence the inferred answer. The question asks: \emph{“What motivated her to change dietary habits?”} with four choices: \textit{(A) family}, \textit{(B) diabetes}, \textit{(C) tooth decay}, and \textit{(D) anemia}. Three frame sets sampled by different methods are shown in Figure~\ref{fig:case}:

\paragraph{Frame set (I)} shows a scattered set of frames showing buildings, family scenes, and unrelated shots. There is no visual evidence related to changing dietary habits. With such weak cues, a plausible guess might be option (A), which is the result obtained by a \textbf{uniform-sampling}. This demonstrates that uniform sampling easily misses events crucial for long-form reasoning.

\begin{figure}
    \centering
    \begin{subfigure}{0.48\columnwidth}
        \centering
        \includegraphics[width=\textwidth]{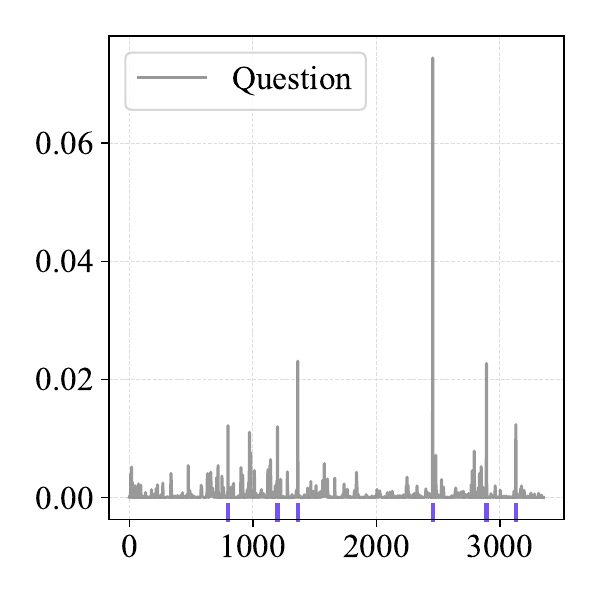}
    \end{subfigure}
    \hfill
    \begin{subfigure}{0.48\columnwidth}
        \centering
        \includegraphics[width=\textwidth]{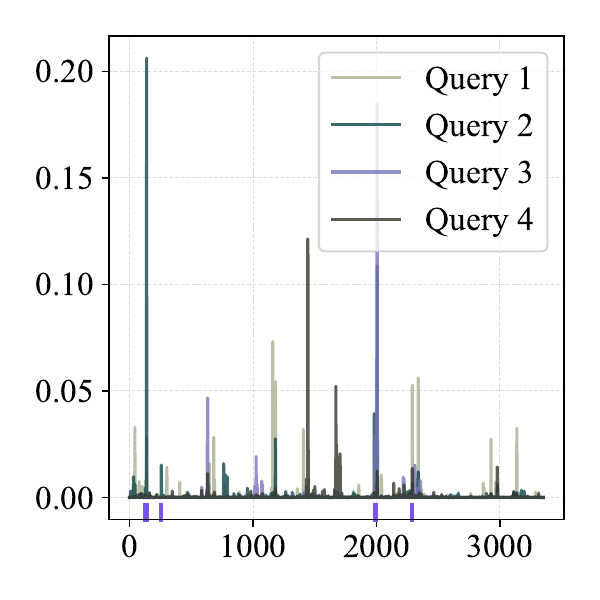}
    \end{subfigure}
    \caption{Distribution of question–frame similarity for top-k (left) and query–frame similarity for \ModelAbbr~(right). Blue markers denote the frames selected by each method.}
    \label{fig:topkvsours}
\end{figure}

\paragraph{Frame set (II)} mostly show people eating, with several duplicates. Although these frames correlate with the noun ``eating'', they lack narrative cues about the dietary change. Given the prevalence of high-calorie food, the model may lean toward (B): the result predicted by \textbf{top-k sampling}. This behavior reflects a limitation of top-k: it tends to over-focus on surface-level lexical matches to the question, instead of exploring alternative reasoning paths. As illustrated in Figure~\ref{fig:topkvsours} (left), the question-frames similarity distribution expresses more noise and lacks of concentration.

\paragraph{Frame set (III)} shows children eating snacks, followed by an interview scene, and then a dentist examining a child with an open mouth. This sequence reveals a clear narrative: the family enjoys snacks, the child develops tooth issues, and they visit the dentist, motivating a change in dietary habits. Thus, an answer of (C) could be inferred: the answer predicted by \textbf{\ModelAbbr}, and also the \textbf{correct answer}.
This is overall processing pipeline of \ModelAbbr~to this question:
\textbf{First}, with a sparse preview and the question as input, \ModelAbbr~reasons out four query concepts (\eg, toothbrush, tooth, doctor, blood test).
\textbf{Next}, using these queries, the CLIP model identifies high-relevance temporal regions of the video. As shown in Figure~\ref{fig:topkvsours} (right), query–frame similarities highlight \textbf{multiple meaningful events} rather than a single narrow peak.
\textbf{Then}, the U-Sampler selects key frames based on the full similarity distribution. Unlike top-k, it does not simply pick the highest scoring frames, the U-Net architecture allows it to prioritize high-scoring \textbf{regions} and maintain temporal diversity. This leads to more coherent narratives and better question alignment.

\section{Conclusion}\label{sec:conclusion}
We introduced \ModelName, a reinforcement learning framework that learns to select key frames for long-form video understanding. The method combines query-driven retrieval, a U-Net based sampler, and joint evolution of the sampler and MLLM to identify informative frames under tight sampling budgets. To support training and evaluation, we built a long-video QA dataset with automatically generated annotations and calibrated difficulty.
Extensive experiments demonstrate that \ModelAbbr~achieves state-of-the-art (SOTA) results on multiple benchmarks while using far fewer frames than dense or heuristic approaches, surpassing base-MLLM by +8 points and the strongest baseline method by +1.1 points on four long-video benchmarks.
Our analysis highlights three key findings: queries provide richer semantic grounding than using the question alone; learned sampling consistently outperforms heuristic strategies; and jointly optimizing the MLLM and sampler is crucial for robust query reasoning and key-frame understanding. These results demonstrate the effectiveness of learning to sample for long-video understanding and offer a scalable direction for future multi-modality systems.

{
    \small
    \bibliographystyle{ieeenat_fullname}
    \bibliography{main}
}

\clearpage
\setcounter{page}{1}
\maketitlesupplementary

\section{More Implementation Details}\label{sup:implementation}
\subsection{Sampler Architecture}
We implement the key-frame sampler with a U-Net with $1\times 3$ convolution layers. This U-Net consists of four down-sampling layers with {32, 64, 128, 256} channels, and four up-sampling layers with reversed channels. To enlarge perception field, the convolution layers are dilated by {1, 2, 4, 8} for the encoders. For the input similarity matrix, query axis is padded to four, denoting a maximum number of four queries (input channels) is processed by the sampler.

\subsection{Training Details}
For \textbf{sampler pre-training}, we use a mini-batch size of 32 to train the sampler for one epoch. We use Adam optimizer with a fixed learning rate of 1e-5. Unlike regular single-label classification tasks, which could directly obtain $\log \textbf{p}(x_{idx})$, our task is a subset selection without replacement. If we simply do $\textbf{p}=\text{softmax}(\textbf{s})$ (\textbf{s} is the output scores of the sampler) and sample $K$ frames, the log probability would be biased, as $\textbf{p}$ converges to $K$ equal peaks with $(N_f-K)$ zeros. This leads to suboptimal loss $K \times \log(1/K)$ while treating all frames \textit{equally}. Due to the \textit{unordered} nature of frame selection, we employ iterative probability calculation where each step conditions on previous selections. Algorithm~\ref{algo:sample} illustrates our approach for proper probability estimation.

\begin{algorithm}
\caption{Probabilistic Sampling Without Replacement}
\label{algo:sample}
\begin{algorithmic}[1]
\Require Scores $\text{scores}$, sample size $K$
\Ensure Selected indices $\text{selected}$, total log probability $\log p_{\text{total}}$

\State $N_f \gets \text{length}(\text{scores})$
\State $\text{remaining} \gets \text{all true vector of size } N_f$
\State $\text{selected} \gets [ ], \text{log\_probs} \gets [ ]$

\For{$k = 1$ \textbf{to} $K$}
    \State Mask unavailable frames in $\text{scores}$ with $-\infty$
    \State $\text{probs} \gets \text{softmax}(\text{masked scores})$
    \State Sample $\text{idx}$ from $\text{Categorical}(\text{probs})$
    \State Append $\log \text{probs}(\text{idx})$ to $\text{log\_probs}$
    \State Append $\text{idx}$ to $\text{selected}$
    \State Set $\text{remaining}[\text{idx}] \gets \text{false}$
\EndFor

\State \Return $\text{selected}$, $\text{sum}(\text{log\_probs})$
\end{algorithmic}
\end{algorithm}

For \textbf{joint RL}, we adopt VERL~\cite{verl} as our training framework and vllm~\cite{vllm} as the inference backend. Learning rate is set to 1e-6 and 1e-5 for the MLLM and sampler, respectively. The training batch size is 32 and group size $G=8$. Following the latest RL training paradigms~\cite{dapo}, we discard KL Divergency regularization, \ie, $kl\_coef=0$. All the experiments are conducted on a single machine with eight H20 GPUs. The following code block includes a minimal RL training script with verl~\cite{verl}:
\begin{tcolorbox}[colback=gray!10, colframe=gray!65!, boxrule=0.5mm]
\small
\begin{lstlisting}[language=bash,breaklines=true]
python3 -m verl.trainer.main_ppo \
data.train_files=hard.parquet \
data.train_batch_size=32 \
data.nframe_init=16 \
data.ntoken_init=32 \
actor_rollout_ref.rollout.agent.nframe_tool=32 \
actor_rollout_ref.rollout.agent.ntoken_tool=256 \
actor_rollout_ref.rollout.agent.sample_method=UNet \
algorithm.adv_estimator=grpo \
algorithm.kl_ctrl.kl_coef=0.0 \
actor_rollout_ref.model.path=Qwen2.5-VL-7B-Instruct \
actor_rollout_ref.actor.optim.lr=1e-6 \
actor_rollout_ref.actor.ppo_mini_batch_size=8 \
actor_rollout_ref.actor.use_kl_loss=False \
actor_rollout_ref.actor.kl_loss_coef=0.0 \
actor_rollout_ref.actor.kl_loss_type=low_var_kl \
actor_rollout_ref.actor.entropy_coeff=0.0 \
actor_rollout_ref.rollout.name=vllm \
actor_rollout_ref.rollout.n=8 \
actor_rollout_ref.rollout.agent.single_response_max_tokens=512 \
trainer.n_gpus_per_node=8 \
trainer.nnodes=1 \
trainer.total_epochs=2
\end{lstlisting}
\end{tcolorbox}

\subsection{Inference Details}
During training (rollout) phase, the inference temperature is set to 1.0 and top-p is set to 0.9, and we use greedy decoding/sampling during evaluation to ensure reproduction. These temperature is set for both the MLLM and sampler.

\section{Dataset Collection Details}\label{sup:dataset}

\paragraph{Step 1: Dense Captioning}
We first collect 2.8k long videos from the Internet, covering diverse domains including movies, documentaries, and sports. To obtain fine-grained textual descriptions, we employ the Gemini-2.5-Flash model to segment each video into semantically coherent scenes and generate captions for each segment. On average, a video is divided into about 20 segments. Gemini is also asked to produce an overall caption summarizing the global context of the entire video. These captions serve as detailed and structured textual annotations that facilitate subsequent question generation.

\paragraph{Step 2: QA Generation}
For each video, we aim to generate cross-event questions that require reasoning over multiple segments. We randomly select pairs of segment captions and merge them with the video’s overall caption. The combined text is then fed to Gemini-2.5-Pro to automatically generate at least one multiple-choice question and its corresponding answer. This process is repeated for different segment combinations, resulting in approximately 20 to 40 QA pairs per video. The questions are designed to require temporal reasoning, comparison, or cause–effect understanding across events rather than within a single segment.

\paragraph{Step 3: QA Filtering}
The automatically generated QA pairs may contain trivial, ambiguous, or incorrect cases. We therefore design a two-stage filtering mechanism to retain only informative and solvable questions.

\textbf{Easy QA filter.} We uniformly sample 16 frames from each video and prompt Qwen2.5-VL-7B-Instruct to answer the question eight times with randomized decoding. If the model answers correctly in at least seven of eight trials, we label the question as too easy and discard it. This step removes questions that can be answered from superficial visual clues or common sense alone.

\textbf{Hard or bad QA filter.} We then sample 256 frames and again query Qwen2.5-VL-7B-Instruct four times. If the model fails to answer correctly even once, we label the question as too difficult or mismatched with the provided answer and discard it. After this stage, about four high-quality QA pairs remain per video, yielding a total of roughly 9.4k QA pairs over 2.8k videos. We denote this dataset as \textbf{\DatasetNameAll}.

\paragraph{Step 4: Difficulty Labeling and Subset Selection}
For reinforcement learning and curriculum evaluation, we further construct a more challenging subset. For each video, we provide all segment captions and the remaining QA pairs to Gemini-2.5-Pro, asking it to rank the relative difficulty of questions within the same video. We then select the top-ranked question as the hardest one. The resulting subset, containing one question per video, is denoted as \textbf{\DatasetNameHard}. This subset emphasizes reasoning-intensive, cross-event questions that best evaluate long-term understanding.

In summary, our dataset construction pipeline combines large-scale automatic generation with fine-grained filtering and difficulty calibration. The resulting dataset provides both scale and diversity, supporting supervised and reinforcement learning of models like \ModelName~that require adaptive reasoning over long-form videos.

\section{More Experiments}

\subsection{Ablation Study}

\paragraph{Reward Design.}
As shown in Table~\ref{tab:abla:reward}, we quantitatively evaluate the effect of the informativeness reward used during joint reinforcement learning (RL) and the difficulty-aware reward employed in the sampler pre-training phase.

A comparison between the first two rows reveals that removing the informativeness reward from RL training results in a slight performance degradation. This suggests that the reward encourages the MLLM to produce more informative and specific queries, as opposed to overly generic ones, thereby enabling the sampler to retrieve more relevant key frames for subsequent video understanding tasks.

The last two rows exhibit a more pronounced performance gap: when the difficulty-aware reward is removed, \ie, when the sampler is trained using a simple binary reward scheme (-1/1), the average accuracy decreases by 3.9 points. This substantial drop underscores the importance of a well-calibrated reward signal in the training process.

\begin{table}[h]
\centering
\caption{Ablation studies on the Informativeness Reward (IR) and Difficulty-aware Reward (DR) with a fixed input budget of 32 frames. LoVi, VLong, and LVB denote the benchmarks LongVideoBench, VideoMME-Long, and LVBench, respectively.}
\label{tab:abla:reward}
\begin{tabular}{cc|cccc}
\toprule
IR & DR & MLVU & LoVi & VLong & LVB \\ \midrule
\cmark                  & \cmark               & 69.3                  & 60.1                  & 54.1                   & 46.4                 \\
\xmark                  & \cmark               & 69.1                  & 59.8                  & 54.1                   & 46.2                 \\ \midrule
 -                  & \cmark                   & 68.2                  & 56.8                  & 52.2                   & 44.3             \\
 -                  & \xmark                   & 62.5                  & 55.4                  & 49.2                   & 38.9                 \\ \bottomrule
\end{tabular}
\end{table}

\paragraph{Sparse Preview of Query Reasoning}
To evaluate the effectiveness of \emph{preview frames} in the first step of \ModelAbbr~inference (query generation), we vary the number of input preview frames and measure their corresponding accuracy. These experiments follow the same configuration as all other experiments, with $K=32$.

\begin{figure}[t]
    \centering
    \includegraphics[width=0.75\linewidth]{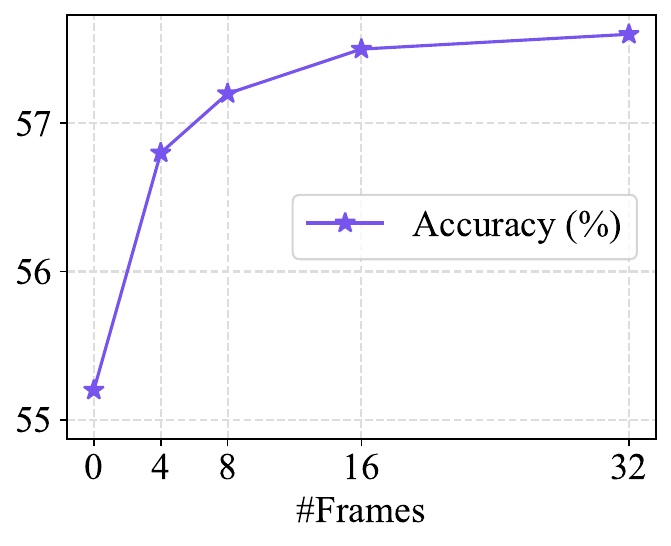}
    \caption{Average accuracy on the four benchmarks under different number of input frames as a preview.}
    \label{fig:previewnframe}
\end{figure}

The results in Figure~\ref{fig:previewnframe} show that without any visual cues ($\#\text{Frames}=0$), the average accuracy drops significantly to approximately 55.2\%. In contrast, even a sparse preview with only four frames achieves substantially higher overall accuracy. This demonstrates the necessity of providing visual context to generate effective queries.

As the number of preview frames increases, a broader trend emerges: doubling the frame budget from 16 to 32 yields only a marginal accuracy gain of 0.1\%, while doubling the computational cost. Therefore, we adopt $K//2=16$ frames at the minimum resolution (32 tokens per frame) as the final configuration.

\subsection{Efficiency Study}
Compared to conventional uniform sampling, our proposed framework introduces several additional inference stages: (i) query reasoning, (ii) similarity matrix computation (FP16), and (iii) frame sampling via U-Net (FP32).
To quantify the resulting overhead, we measure the time consumption by iterating over 1,000 samples through these steps.
The results in Table~\ref{tab:time} show that both similarity matrix generation and the U-Net forward pass incur negligible time with each accounting for less than 1\% of the total cost.

Although the overall time increases by 30\% compared to vanilla uniform sampling, our question-answering (QA) step is 10\% faster, despite using the same number of input tokens as in uniform sampling.
This speedup can be attributed to prefix caching in the vLLM backend during the query reasoning phase, which reduces computation in the subsequent answer generation phase due to cache hits.
Quantitatively, the prefix cache hit rates for our pipeline and uniform sampling are 17.2\% and 11.8\%, respectively.

\begin{table}[h]
\centering
\caption{Average time (s) cost by the four inference steps on 1,000 samples.}
\label{tab:time}
\begin{tabular}{ccccc|c}
\toprule
Query & CLIP & UNet & QA & ALL & Uniform \\ \midrule
2.809                  & 0.029                  & 0.038                   & 3.353              & 6.226 & 3.723   \\ \bottomrule
\end{tabular}
\end{table}

\subsection{Reward Visualization}

\begin{figure}
    \centering
    \includegraphics[width=\linewidth]{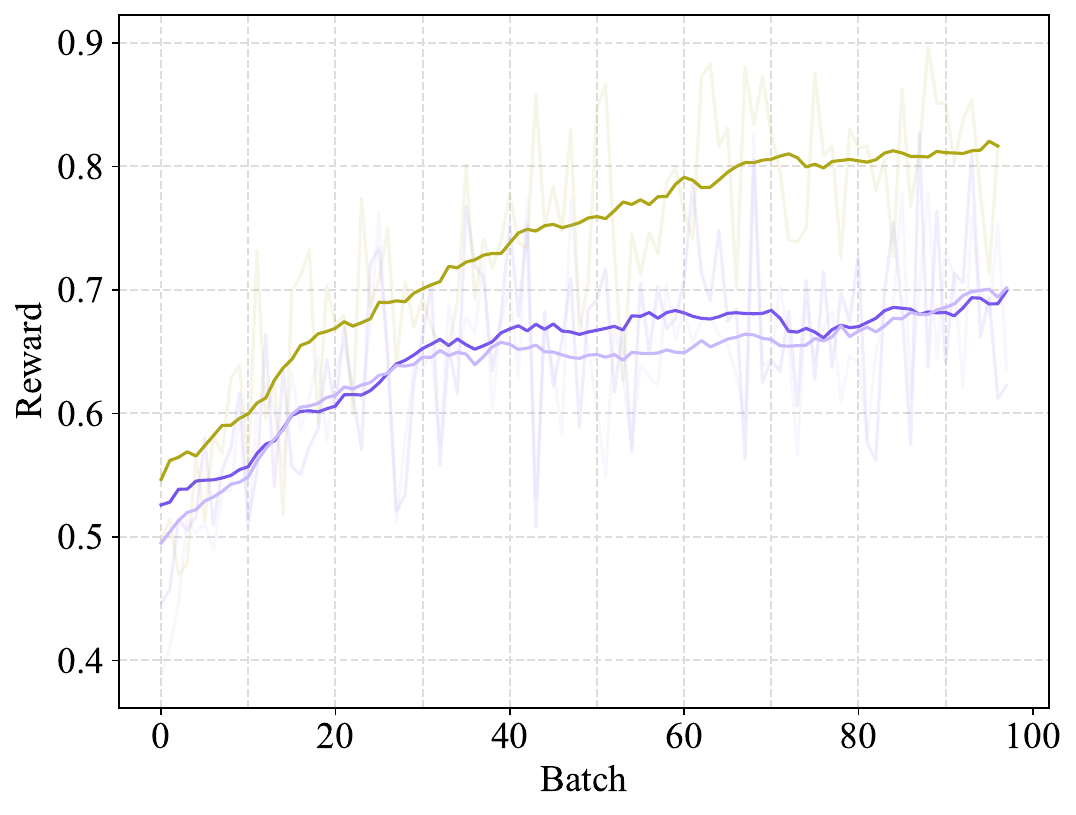}
    \caption{Smoothed reward curves across different training configurations.}
    \label{fig:reward}
\end{figure}

We present the training reward curves in Figure~\ref{fig:reward}, which compares three experimental settings: training on the full \DatasetNameAll~dataset (yellow), training on the more challenging \DatasetNameHard~dataset with a pre-trained sampler (violet), and training on \DatasetNameHard~without pre-training (light violet).

As shown in the figure, the reward on the full dataset increases rapidly during the initial training phase. In contrast, progress on the harder dataset is more gradual, reflecting its greater difficulty. While the simpler dataset saturates quickly, the increased complexity of the harder dataset appears to better leverage the model's capacity, ultimately yielding a 2.5\% overall performance improvement on our benchmarks.

Comparing the two \DatasetNameHard~conditions reveals that initializing with a pre-trained sampler significantly stabilizes the early stages of training. This improved stability contributes to enhanced final performance, demonstrating the value of a well-initialized sampler for challenging learning tasks.

\newpage
\section{Prompts}
This section details the prompt templates used for query generation and question answering in our framework. The prompt design is motivated by several key considerations: (1) \textit{Explicit role definition} through system prompts establishes clear model behavior; (2) \textit{Structured constraints} ensure outputs conform to downstream processing requirements; (3) \textit{Visual grounding} prioritizes observable content over abstract reasoning; and (4) \textit{Format standardization} enables reliable parsing and integration with our retrieval pipeline. The prompts below implement these principles through precise instruction phrasing, response formatting rules, and example demonstrations.

\begin{tcolorbox}[
    colback=gray!5,
    colframe=gray!80!black,
    boxrule=0.5mm,
    arc=2mm,
    fonttitle=\bfseries,
    title=System Prompt
]
You are an expert video analyst. Strictly follow the user's instructions.
\end{tcolorbox}

\begin{tcolorbox}[
    colback=gray!5,
    colframe=gray!80!black,
    boxrule=0.5mm,
    arc=2mm,
    fonttitle=\bfseries,
    title=Query Generation Prompt
]
\textbf{Task: Visual Descriptive Queries Generation}

Generate \textbf{<=4} distinct and visually grounded descriptive queries based on the question and video. These queries will be sent to a CLIP model to retrieve relevant frames for answering the question.

\textbf{Response Format:} List queries as an array within \texttt{<queries></queries>} tags.

\textbf{Strict Guidelines:}
\begin{itemize}[leftmargin=*,noitemsep,topsep=2pt]
    \item \textbf{Visually grounded:} Describe only observable elements
    \item \textbf{Question grounded:} Focus on visual cues that help answer the question
    \item \textbf{Diverse:} Each query must describe a distinct visual perspective
    \item \textbf{Length:} <=4 queries, each under 20 words
    \item \textbf{Format:} Start each query with ``An image of ...''
    \item \textbf{Priority:} Sort queries from most to least important for answering
\end{itemize}

\textbf{Example Response:}
\begin{verbatim}
<queries>
query 1: An image of ...
query 2: An image of ...
...
</queries>
\end{verbatim}

\textbf{Input:}
\begin{verbatim}
Initial Video: <video>
Question and Options: ...
\end{verbatim}
\end{tcolorbox}

\begin{tcolorbox}[
    colback=gray!5,
    colframe=gray!80!black,
    boxrule=0.5mm,
    arc=2mm,
    fonttitle=\bfseries,
    title=Question Answering Prompt
]
\textbf{Task: Question Answering}

Focus on the \textbf{Current Video}, directly answer the multiple-choice question without any extra text (a single upper-case letter of the option).

\textbf{Response Format:} Place final answer within \texttt{<answer></answer>} tags.

\textbf{Example Response:}
\begin{verbatim}
<answer>A</answer>
\end{verbatim}

\textbf{Input:}
\begin{verbatim}
Current Video: <video>
Question and Options: ...
\end{verbatim}
\end{tcolorbox}

\end{document}